\documentclass{article}

\usepackage[final]{nips_2017}
\usepackage[ruled]{algorithm2e}
\usepackage{graphicx}
\usepackage{float}
\usepackage[mathscr]{euscript}

\usepackage[utf8]{inputenc} 
\usepackage[T1]{fontenc}    
\usepackage{hyperref}       
\usepackage{url}            
\usepackage{booktabs}       
\usepackage{amsfonts}       
\usepackage{nicefrac}       
\usepackage{microtype}      

\title{Learnings Options End-to-End for Continuous Action Tasks}

\author{Martin Klissarov, Pierre-Luc Bacon, Jean Harb, Doina Precup\\
Reasoning and Learning Lab,\\
McGill University \\
{\tt \{mklissa,pbacon,jharb,dprecup\}@cs.mcgill.ca}
}

\usepackage{amssymb,amsmath}
\usepackage{amsthm}
\usepackage{bm}
\def\given{\middle\vert}

\def\expectation{\mathbb{E}}

\def\prob{P}
\def\defeq{\dot=}
\newcommand{\deriv}[2][]{\frac{\partial#1}{\partial#2}}

\def\State{S}
\def\states{\mathscr{S}}

\def\Action{A}

\def\actions{\mathscr{A}}
\def\option{o}
\def\opt{\option}
\def\Option{O}

\begin{document}

\maketitle

\begin{abstract}
  We present new results on learning temporally extended actions for continuous tasks, using the options framework (\cite{SuttonPrecupSingh1999}, \cite{Precup2000}). In order to achieve this goal we work with the option-critic architecture~(\cite{Bacon2017}) using a deliberation cost and train it with proximal policy optimization~(\cite{SchulmanWDRK17}) instead of vanilla policy gradient. Results on Mujoco domains are promising, but lead to interesting questions about \textit{when} a given option should be used, an issue directly connected to the use of initiation sets.

\end{abstract}
\vspace*{-8pt}

\section{Introduction}
\vspace*{-8pt}

The options framework (\cite{SuttonPrecupSingh1999}, \cite{Precup2000}) allows a reinforcement learning agent to represent, learn and plan with temporally extended actions. These temporally extended actions consist of a set of internal policies, termination conditions and sometimes initiation sets that allow controlling the number of choices available to an agent. Given a set of options, the agent will learn a policy over options, which is typically viewed as executing in a call-and-return fashion: once this policy chooses an option, the option will execute until it terminates, then the policy over options will make a new choice. Learning options is beneficial as it leads to specialization in the state space, and therefore to potentially reduced complexity in terms of the internal policies of the options. 
The option-critic architecture (\cite{Bacon2017}) provides an agent with an end-to-end algorithm to learn options in order to maximize the expected discounted return, by relying on ideas akin to actor-critic methods.
In this work, we exploit the option-critic architecture by combining it to a recent algorithm, Proximal Policy Optimization (PPO)~(\cite{SchulmanWDRK17}), which is very well suited for continuous control tasks and has shown better sample complexity in empirical comparisons. We present results of our approach on a set of environments from the Mujoco framework; our results are consistent with published evaluations which show that learning options provides increased performance, better interpretability and faster learning.  
\vspace*{-8pt}

\section{Background}
\label{gen_inst}
\vspace*{-8pt}

A Markov Decision Process $\mathcal{M}$ is a tuple $ \defeq (\mathcal{S}, \mathcal{A}, \gamma, r, P)$ with $\mathcal{S}$ the state set, $\mathcal{A}$ the action set and the scalar $\gamma\in [0,1)$ the discount factor. The reward function maps states and actions to a scalar reward $r : \states \times \actions \rightarrow Dist(\mathbb{R})$ and the transition matrix $P: \states \times \actions \to Dist(\states)$ specifies the environment's dynamics.
A policy $\pi$ is a set of probability distributions over actions conditioned on states $\pi$: $\states \to \actions$. For a given policy, the value function $V_\pi(s) \defeq \expectation_\pi\left[ \sum_{t=0} \gamma^t r(S_t, A_t) \given S_0 = s\right]$ defines the expected return obtained by following $\pi$. $V_\pi$ satisfies the Bellman equations : $ V_\pi(s) = \sum_{a} \pi\left(a \given s\right)\left( r(s, a) + \gamma \sum_{s'} \prob\left(s' \given s, a\right) V_\pi(s')\right)$.

The policy gradient theorem (\cite{Sutton1999}) provides the gradient of a parametrized stochastic policy $\pi_\theta$ with respect to the expected discounted return from an initial state distribution $d_0 \in \text{dist}(\states)$. For simplicity, we write the policy as $\pi$, making its parametrization ($\theta$) implicit.
\begin{align*}
    \deriv[L(\theta)]{\theta} = \sum_{s} d(s;\theta) \sum_{a} \deriv[\pi\left(a \given s\right)]{\theta}Q_{\pi}(s, a)
\end{align*} 
where  $d(s;\theta) = \sum_{s_0} d(s_0) \sum_{t=0}^{\infty} \gamma^t P_{\pi}(S_t = s | S_0 = s_0)$
is a weighting of states along the trajectories generated by $\pi$ and passing through $s$. 
Using the log-likelihood trick (\cite{Williams1992}),
\begin{align*}
    \deriv[L(\theta)]{\theta} = \expectation\left[ \deriv[\log \pi\left(A_t \given S_t\right)]{\theta} A^{\pi}(S_t, A_t)  \right]
\end{align*}
where $A^{\pi}(S_t, A_t) = Q_{\pi}(S_t, A_t) - V_{\pi}(S_t)$ is the advantage function. The term $V_{\pi}(s_t)$ acts as a \textit{baseline} (\cite{Williams1992,Sutton1999}) which reduces the variance of the resulting
estimator. 
\vspace*{-8pt}

\subsection{Trust region methods and Proximal Policy Optimization (PPO)}
\vspace*{-8pt}

Trust region methods, and in particular the TRPO algorithm (\cite{SchulmanLMJA15}), are second-order methods that maximize a surrogate objective subject to a constraint.
TRPO has proven useful for continuous control, but it can be computationally expensive and doesn't allow for parameter sharing.

Proximal Policy Optimiation (PPO) achieves the same level of reliability and performance as TRPO while being a first-order method. To do so, it uses an objective with clipped probability ratios, preventing an excessive shift in the probability distribution between updates. This clipping also allows for multiple epochs of minibatch updates on a single sampled trajectory. The clipped surrogate objective is:
\begin{align*}
  \deriv[L(\theta)^{\text{PPO}}]{\theta}  = \expectation\left[\deriv[]{\theta} \min( \rho_t(\theta) A^{\pi}(S_t, A_t), \text{clip}(\rho_t(\theta), 1 - \epsilon, 1+ \epsilon) A^{\pi}(S_t, A_t))\right]
\end{align*}
where $\rho_t(\theta) = \frac{\pi(A_t | S_t)}{\pi_{{old}}(A_t | S_t)}$ is the importance sampling ratio. The authors use the Generalized Advantage Estimation (\cite{SchulmanMLJA15}) to calculate the advantage function $A^{\pi}(S_t, A_t)$.
\vspace*{-8pt}

\subsection{Option-Critic}
\vspace*{-8pt}

The option-critic architecture (\cite{Bacon2017}) is a gradient-based approach for learning intra-option policies as well termination conditions, assuming that all options are available at every state. Moreover, the parameters of the intra-option policies ($\theta_{\pi}$) and the termination function ($\theta_{\beta}$)  are assumed to be independent. The intra-option policy gradient is as follows:
\begin{align*}
    \deriv[L(\theta)]{\theta_\pi} =  \expectation\left[ \deriv[\log \pi\left(\Action_t \given \State_t,\Option_t\right)]{\theta_\pi} Q_\pi(\State_t,\Option_t, \Action_t)\right] \enspace 
\end{align*}
where a baseline (i.e. the above state-option value function $Q_\pi $  parametrized by $\theta_w$) is generally added. However, if the options are learned to optimize returns, in the long run, they will tend to disappear since any MDP can be solved optimally using primitive actions. To avoid this problem, \cite{Harb2018} use the bounded rationality framework (\cite{Simon1969}) and introduce a deliberation cost ($\eta$), interpreted as  a margin of how much better an option should be than the current option in order to replace it. The termination gradient then takes the following form:
\begin{align*}
    \deriv[L(\theta)]{\theta_\beta} = \expectation\left[ -\deriv[\beta(\State_t, \Option_t)]{\theta_\beta} (A^{\pi \beta}(\State_t,\Option_t) + \eta) \right] 
\end{align*}
where $A^{\pi \beta}(\State_t,\Option_t) = Q_\pi(\State_t,\Option_t) - V_\pi(\State_t)$ is the termination advantage function and stems directly from the derivation of the gradient. 
\vspace*{-8pt}

\section{Algorithm}
\vspace*{-8pt}

We introduce the Proximal Policy Option-Critic (PPOC) algorithm which, just like PPO, works in two stages. In the first stage, the agent collects trajectories of different options and computes the advantage functions using Monte-Carlo returns. We then proceed to the optimization stage where, for K optimizer iterations, we choose M tuples and apply the gradients. We also chose to use a stochastic policy over options, parameterized by an independent vector $\theta_{\mu}$ (as opposed to $\epsilon$-greedy) which we learned under the same policy gradient approach. 

\begin{algorithm}[ht]
\DontPrintSemicolon
\SetAlgoLined
\For{iteration=1,2,....}{
    $c_t \leftarrow 0$\;
    $s_t \leftarrow s_0$\;
    Choose $\option_t$ with a softmax policy over options $\mu(\option_t|s_t)$ \\
    \Repeat{T timesteps} {
        Choose $a_t$ according to $\pi(a_t|s_t)$ \;
        Take action $a_t$ in $s_t$, observe $s_{t+1}$, $r_t$\;
        $\hat r_t = r_t - c_t$\;
        \uIf{$\beta$ terminates in $s_{t+1}$}{
        choose new $\option_{t+1}$ according to softmax $\mu(\option_{t+1}|s_{t+1})$\;
        $c_t = \eta$}
        \Else{$c_t = 0$}
    }
    
    Compute the advantage estimates for each timestep; \\
    \For{$\option$=$\option_1$,$\option_2$,....}{
        $\theta_{old} \leftarrow \theta$\;
        \For{K optimizer iterations with minibatches M}{
            $\theta_{\pi}  \leftarrow \theta_{\pi} + \alpha_{\theta_{\pi}} \deriv[L_t(\theta)^{\text{PPO}}]{\theta_\pi} $ \;
            
            $\theta_\beta \leftarrow \theta_\beta - \alpha_{\theta_\beta} \deriv[\beta(s_t)]{\theta_\beta} \left( A(s_t,o_t) + \eta \right)\;$\;
            
            $\theta_{\mu}  \leftarrow \theta_{\mu} + \alpha_{\theta_{\mu}} \deriv[\log\mu(\option_t|s_t)]{\theta_\mu}  A(s_t,\opt_t) $ \;
            
            $\theta_w \leftarrow \theta_w - \alpha_{\theta_w} \deriv[(G_t - Q_\pi(s_t,\option_t))^2]{\theta_w}$\;
            
        }
    }
    
}

\caption{Proximal Policy Option Critic (PPOC)}
\end{algorithm}
\section{Experiments}
We performed experiments on locomotion tasks available on OpenAI's Gym (\cite{BrockmanCPSSTZ16}) using the Mujoco simulator (\cite{Todorov2012MuJoCoAP}).  We aim to assess the following: (1) whether the use of options can increase the speed of learning as well as the final performance, (2) the interpretability of the resulting options.

In our experiments, we used as input the vectors defining joint angles, joint velocities, and coordinates of the center of mass. We used two separate networks with 64 hidden units per layer, each containing two layers.\footnote{The code, as well as the values for the hyperparameters, are available here: \url{https://github.com/mklissa/PPOC}} For all the layers we used tanh non-linearity, except for the output which was linear for of value functions and intra-option policies, sigmoid for the termination probability and softmax for the policy over options. The first network was used to output the policy over options $\mu(\option | s)$ and the intra-option policies $\pi(a | s)$, while the second network was used to output the value functions $Q_\pi(s,\option)$ and the termination probabilities $\beta(s)$. The log-standard deviations were parameterized by a vector independent of the input state. 
We used the exact same hyper-parameters as mentioned in \cite{SchulmanWDRK17}, except for the optimizer mini-batch size which was divided by the number of options. We proceeded so in order to avoid training more samples per iteration with the options framework, thus enabling a fair comparison between options and primitive actions. In the case of options, we also divide the reward by 10 to reduce the scale of the value functions, and therefore the termination probability gradient, making it more stable.  We didn't proceed to any hyper-parameters search to improve the results. Our experiments exclusively investigate the merits of using two options and compare the results to the case of primitive actions (no options).

In addition to the classic Mujoco environments, we ran agents in an environment called HopperIceBlock-v0.\footnote{HopperIceBlock-v0 is based on \cite{henderson2017multitask} and is avaialable here: \url{https://github.com/mklissa/gym-extensions}} This environment contained a more explicit compositionality than the original Mujoco environments available on OpenAI's Gym, which simply require  to learn a gait that maximizes speed in a direction. We used Hopper-v1 as a starting point and added some obstacles in the agent's path: solid slippery blocks. The agent had to learn to pass them either by jumping completely over them or by sliding on their surface.

\begin{figure*}[ht]
  \centering

  \includegraphics[width=\textwidth]{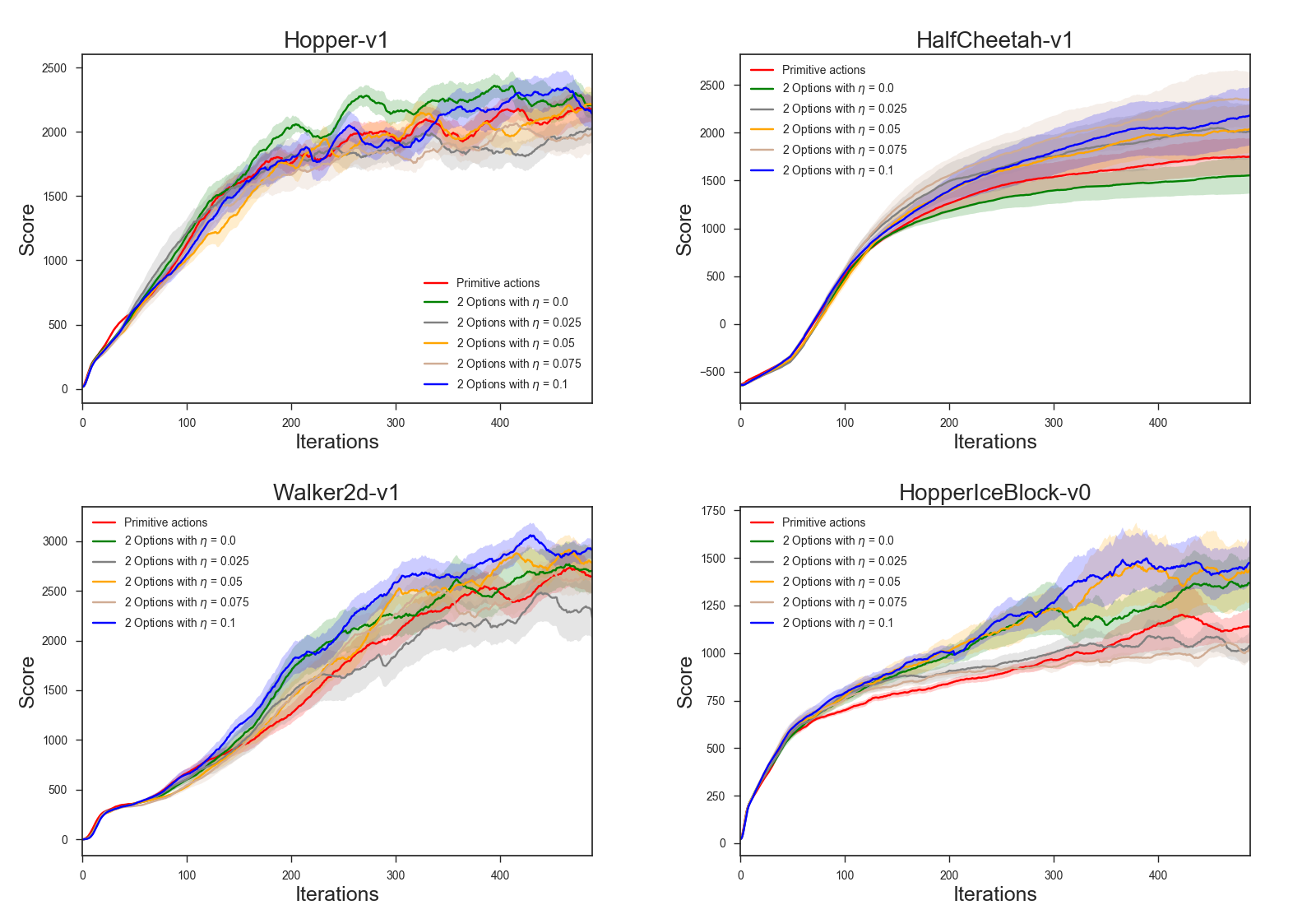}
  \caption{Results in Mujocco using 12 different random seeds for a total of 1 million steps (each iteration is 2000 steps)}
\end{figure*}

The results are summarized in Fig.1. As expected, using options with a deliberation cost yields better results and faster learning on most environments. It is interesting to note that the increase in performance is not directly proportional to the value of $\eta$. This is due to the different scales of the average returns across  environments, as well as during the course of learning. In the current formulation of the deliberation cost, its value is a hyperparameter that has to be set. It would be useful to explore the possibility of working with a learned value instead of a constant. This is left as future work.

The results that stand out the most are the one on the customized environment. More importantly, the success threshold for the environment is around 1200 points, under that level the agent actually doesn't learn to pass the iceblock and continue its gait.  So, the agent using options is the only one solving this environment. This also led us to investigate how the options are used in this environment as opposed to the classic Mujoco environments.\footnote{Videos from the environments are available at \url{https://www.youtube.com/watch?v=XI_txkRnKjU}} In HopperIceBlock-v0, the interpretability of the options is obvious and greatly helps the performance: one option is used to hop when there is no iceblock nearby, but then when passing over the iceblock, both options are used to complete the specific task. 
In the case of the classic Mujoco environments, one option is used to gain momentum at the start of the episode and is never used thereafter. Even if the agents using options outperform the agents with primitive actions on classic environments, we can only truly see the benefits of a hierarchical framework when used in the appropriate environment. 

\section{Conclusion}

Our experiments  demonstrate that it is possible to learn options in an end-to-end manner using deep networks on continuous actions environments, and to the best of our knowledge this is the first work to do so. Our results also suggest that the increase in performance is not directly linked to the deliberation cost, which is problematic as it leaves us with the task of finding the right value. For the options framework to be truly end-to-end it would be necessary to learn a value of $\eta$.
More importantly, we have seen that the increase in performance is related to the compositionality of the environment. In the classic Mujoco environments, using options is not as beneficial as using them in a customized environment with a more obvious division in the state-space. This leads to the following question: when should we be using options? This question also points to a fundamental problem in the current options framework: it is necessary for us to manually specify the number of available options. How should one decide on this number? As stated in \cite{Bacon2017}, one way to answer this question would be to reintroduce the notion of initiation sets in the option-critic architecture.

\bibliographystyle{named}
\bibliography{references}

\clearpage

\end{document}